\documentclass[runningheads]{llncs}

\usepackage{eccv}

\usepackage{eccvabbrv}

\usepackage{graphicx}
\usepackage{array}
\usepackage{booktabs}
\usepackage{amsmath}
\usepackage{amssymb}
\usepackage{longtable}
\usepackage{placeins}
\usepackage{float}
\usepackage{pgfplots}
\pgfplotsset{compat=1.18}

\definecolor{wsiviolet}{HTML}{7C3AED}
\definecolor{wsimuted}{HTML}{6B7280}
\definecolor{wsigrid}{HTML}{EDEFF3}

\usepackage[accsupp]{axessibility}  

\usepackage{hyperref}
\hypersetup{
  pdftitle={DistillPath: An Efficient 22M Distilled Pathology Encoder Approaching Large Foundation Model Performance},
  pdfauthor={Ramon Kaspar, Andrey Ignatov, Valentina Boeva}
}

\usepackage{orcidlink}

\begin{document}

\title{DistillPath: An Efficient 22M Distilled Pathology Encoder Approaching Large Foundation Model Performance\texorpdfstring{\thanks{Accepted at the ECCV 2026 Workshop on Medical Foundation Models and Benchmarks (MedFM-Bench).}}{}}

\titlerunning{DistillPath: 22M ViT-S/16 Distillation}

\author{Ramon Kaspar\orcidlink{0009-0005-9431-2017} \and
Andrey Ignatov\orcidlink{0000-0003-4205-8748} \and
Valentina Boeva\orcidlink{0000-0002-4382-7185}}

\authorrunning{R.~Kaspar et al.}

\institute{Department of Computer Science, ETH Z\"urich, Switzerland\\
\email{kasparr@ethz.ch, valentina.boeva@inf.ethz.ch}}

\maketitle

\begin{abstract}
  Many high-performing pathology tile encoders are now foundation models with hundreds of millions to over a billion parameters. Encoding and storing the thousands of tiles in each whole-slide image with such models is costly on commodity hardware, so compact encoders that retain useful downstream performance are a valuable alternative. We present DistillPath-KS16, which starts from the existing 22M kaiko ViT-S/16 encoder and improves it by distilling from released pathology encoders used as frozen teachers. The recipe reads only the teachers' final class and patch tokens and trains on 6{,}000 public slides, needing neither their DINO nor iBOT pretraining heads nor a billion-tile corpus, so it applies to any released encoder that exposes backbone tokens. We distill four teachers spanning 86M to 1.1B parameters into the same student. Every variant improves the kaiko baseline on all three benchmarks we use, EVA, HEST, and PLISM, and the strongest teacher is task-dependent. On the seven-task EVA mean, DistillPath-KS16-Virchow2 reaches $0.795$, within $0.015$ points of Virchow2, the top-scoring model in our evaluation, at about $29\times$ fewer parameters; it also scores above H0-mini and GPFM on this aggregate metric, though that advantage is task-concentrated rather than uniform. Because it remains a 22M ViT-S/16 with 384-dimensional features, DistillPath-KS16 runs more than $25\times$ faster than Virchow2. Code is available at \url{https://github.com/RamonKaspar/DistillPath}, and released model weights are available at \url{https://huggingface.co/collections/RamonK/distillpath}.
\end{abstract}

\section{Introduction}
\label{sec:intro}

Computational pathology pipelines often split large whole-slide images into thousands of tiles and encode every tile with a pretrained encoder~\cite{clam}. Many high-performing recent tile encoders are pathology foundation models (FMs) with hundreds of millions to more than a billion parameters, such as UNI~\cite{uni,uni2}, Virchow2~\cite{virchow2}, H-optimus-0~\cite{hoptimus0}, and Prov-GigaPath~\cite{provgigapath}. This scale is costly because whole-slide inference applies the encoder thousands of times per slide and downstream pipelines often store every tile embedding. A compact encoder that preserves the downstream performance can therefore reduce inference time, memory pressure, and feature-storage cost.

Knowledge distillation can train a compact encoder from a large teacher without repeating large-scale self-supervised pretraining~\cite{hinton,duval2023}. In pathology, H0-mini distills H-optimus-0 into a ViT-Base through the teacher's DINO~\cite{dino} and iBOT~\cite{ibot} heads~\cite{h0mini}, Virchow2G-Mini distills Virchow2G into a ViT-Small on a billion tiles with a DINOv2-style head-based recipe~\cite{virchow2}, and GPFM matches the backbone features of several expert encoders while pretraining a ViT-Large from scratch with DINOv2~\cite{gpfm}. The first two need the teacher's pretraining heads, which many released encoders do not provide, while the third folds feature matching into a larger, expensive self-supervised run. This motivates a recipe that uses only frozen backbone outputs to transfer that signal into a compact, already pathology-pretrained tile encoder.

We introduce DistillPath-KS16, a family of 22M kaiko ViT-S/16~\cite{kaiko} pathology encoders distilled from released teacher encoders. The recipe reads only the teacher's final class and patch tokens. It aligns the class token with a cosine loss and a relational loss, and aligns the patch tokens with a cosine loss. We apply this recipe to four teachers that span a wide range of sizes and training data: H0-mini (86M), Virchow2 (632M), UNI2-h (681M), and H-optimus-0 (1.1B). Each is distilled into the same student on 6{,}000 public TCGA slides~\cite{tcga}.

\begin{figure}[t]
  \centering
  \includegraphics[width=\textwidth]{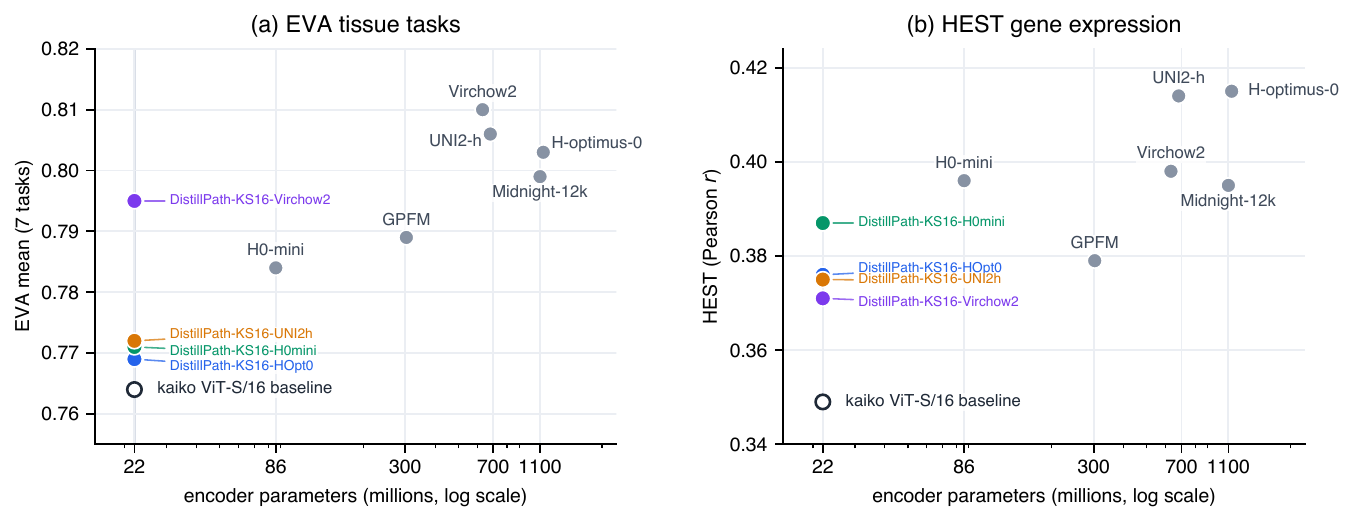}
  \caption{\textbf{Downstream score versus encoder size.} Four teachers are distilled into the same 22M kaiko ViT-S/16. DistillPath-KS16-Virchow2 improves the EVA mean from $0.764$ to $0.795$, within $0.015$ of Virchow2 using about $29\times$ fewer parameters. (a) EVA mean over seven tasks; (b) mean HEST Pearson $r$ over nine tasks.}
  \label{fig:pareto}
\end{figure}

Figure~\ref{fig:pareto} summarizes the main result. Every one of the four distilled variants improves the kaiko baseline on all three benchmarks. The best variant, DistillPath-KS16-Virchow2, reaches $0.795$ on the seven-task EVA mean and also scores above H0-mini ($0.784$) and GPFM ($0.789$) on this aggregate metric, although that advantage is concentrated in a few tasks rather than uniform, as detailed in Section~\ref{sec:results}. Which teacher is best depends on the target task: the teacher's own score is not sufficient to predict the student's, and the largest teacher, H-optimus-0, gives the lowest-EVA variant while the smallest, H0-mini, gives the strongest HEST student.

This paper makes three contributions. First, we present DistillPath-KS16, an efficient 22M distilled pathology encoder family based on kaiko ViT-S/16. Second, our backbone-token distillation recipe reads only frozen teacher class and patch tokens, uses no teacher pretraining heads, and trains on public slides, so it can be applied to any released encoder that exposes backbone tokens. Third, we quantify the performance-efficiency tradeoff: the best variant reaches $0.795$ EVA while running more than $25\times$ faster than Virchow2 in our encoder-forward benchmark.

\section{Related Work}

\subsection{Foundation models for computational pathology}

Earlier pathology pipelines commonly used ImageNet-pretrained encoders~\cite{clam}; later work learned representations directly from pathology images. CTransPath uses a Swin Transformer with a convolutional stem~\cite{ctranspath}, while RetCCL uses a convolutional encoder~\cite{retccl}. Recent FMs mainly use ViT~\cite{vit} backbones with self-supervised objectives such as DINOv2 and masked-image-modeling variants like iBOT~\cite{ibot,dinov2,phikon}. These include Phikon and Phikon-v2~\cite{phikon,phikonv2}, UNI and UNI2-h~\cite{uni,uni2}, Virchow2~\cite{virchow2}, H-optimus-0 and H-optimus-1~\cite{hoptimus0,hoptimus1}, Prov-GigaPath~\cite{provgigapath}, and the kaiko models~\cite{kaiko,midnight}. Their sizes range from 22M ViT-Small models to ViT-giant models with more than one billion parameters. PLUTO-4S reaches the same 22M size through direct self-supervised pretraining on a large proprietary corpus rather than distillation from a released teacher, and its weights are not public, so we do not include it as an experimental baseline~\cite{pluto4}. Size is not the only factor: Midnight reaches competitive results with much less training data~\cite{midnight}, and Virchow2 studies the role of data diversity and pathology-specific training changes~\cite{virchow2}.

\subsection{Distillation of pathology foundation models}

Knowledge distillation trains a student to match a fixed teacher~\cite{hinton}. Existing pathology distillation approaches differ in what they match and how. H0-mini follows the DINOv2 distillation recipe of Duval \etal~\cite{duval2023,h0mini} and passes the class and patch tokens through the teacher's pretrained DINO and iBOT heads. Virchow2G-Mini is the most directly comparable prior model we identify, since it also distills into a 22M ViT-Small; it uses a DINOv2-style recipe to distill Virchow2G on one billion tiles with a large compute budget~\cite{virchow2}. Both therefore require the teacher's pretraining heads, which are unavailable for many released encoders. GPFM takes a different route: it matches backbone features from UNI, Phikon, and CONCH~\cite{conch} while pretraining a ViT-Large from scratch with DINOv2, and its expert loss aligns class tokens with cosine distance and patch tokens with cosine distance plus a pointwise penalty~\cite{gpfm}. Our recipe also uses only the frozen backbone, like the GPFM expert loss, but it is the sole training objective for a small student that is already pretrained, rather than an auxiliary loss inside a full self-supervised run, so it stays usable with any released teacher that exposes backbone tokens.

\subsection{Feature-distillation objectives}

Pointwise feature distillation aligns each student output with its corresponding teacher output. A relational objective instead matches the geometry of a batch: RKD compares normalized pairwise distances and triplet angles between samples~\cite{rkd}. Relational matching transfers representation geometry without requiring equal teacher and student dimensions. On the class token, our recipe uses both a pointwise cosine loss and RKD, because they constrain different properties of the representation; on the patch tokens, it uses a pointwise cosine loss.

\section{Method}
\label{sec:method}

\begin{figure}[t]
  \centering
  \includegraphics[width=\textwidth]{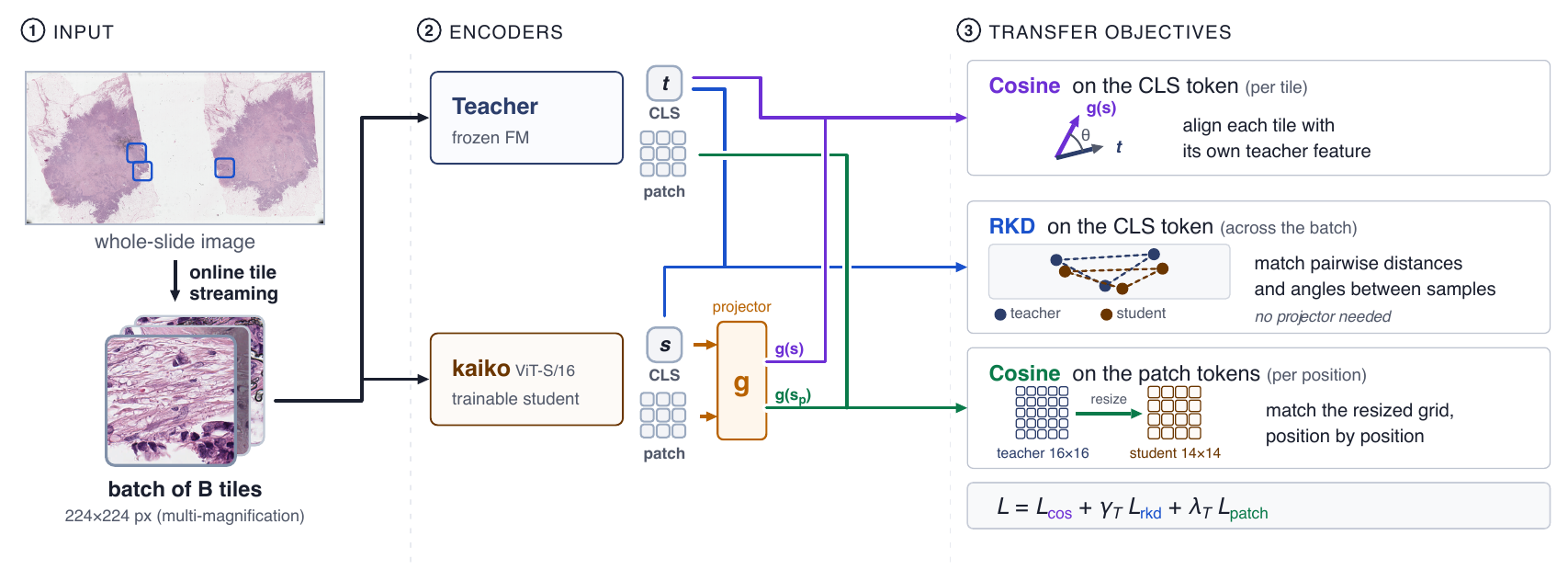}
  \caption{\textbf{Method overview.} Tiles are streamed online from whole-slide images and encoded by a frozen teacher and the trainable kaiko ViT-S/16 student. We read only the final class and patch tokens. The class token is supervised by a cosine loss (per tile) and a relational RKD loss (across the batch). The patch tokens are supervised by a cosine loss after the teacher grid is resized to the student grid. A shared projector $g$ maps student features to the teacher dimension for the cosine losses.}
  \label{fig:method}
\end{figure}

The same image tile is passed to a frozen teacher and a trainable ViT student, and we supervise the student with the teacher backbone's final class and patch tokens, without the teacher's pretraining heads. For the pointwise losses, a trainable projector maps student features to the teacher dimension; RKD compares relations within each feature space and needs no projector. After training, the teacher and projector are discarded, so inference uses the student backbone alone. Figure~\ref{fig:method} gives an overview.

\subsection{Distillation objectives}

For an input tile, the teacher and student return token sequences. Some teachers also return register tokens, which we ignore~\cite{registers}. We denote the student and teacher class tokens by $s$ and $t$, and their patch tokens by $s_p$ and $t_p$. The student dimension is $d_s=384$, and the teacher dimension $d_t$ depends on the teacher.

\paragraph{Pointwise class-token loss.} Following the class-token term of the GPFM expert feature loss~\cite{gpfm}, a projector $g$ maps the student token to the teacher dimension, and we align each projected student class token with its teacher class token using cosine distance,
\begin{equation}
  \mathcal{L}_{\text{cos}} = 1 - \cos\!\big(g(s),\, t\big),
\end{equation}
averaged over the batch. The DINO-style projector is Linear$(384,2048)$--GELU--Linear$(2048,2048)$--GELU--Linear$(2048,256)$, followed by $\ell_2$ normalization and a bias-free weight-normalized Linear$(256,d_t)$ output layer~\cite{dino}. MLP linear weights use truncated-normal initialization with standard deviation $0.02$, and biases in the MLP are initialized to zero. The projector is applied once to the full student token sequence, so the class- and patch-token losses share the same projector.

\paragraph{Relational class-token loss.} RKD compares relations within a batch instead of matching each feature directly~\cite{rkd}. The distance term compares pairwise Euclidean distances after normalizing each distance matrix by its mean off-diagonal value, $\mu_s$ or $\mu_t$. The angle term compares triplets around each anchor embedding. Both use a Huber penalty $\ell_\delta$ with $\delta=1$,
\begin{align}
  d_z(i,j) &= \tfrac{\lVert z_i-z_j\rVert}{\mu_z}, \qquad z\in\{s,t\}, \notag\\
  \mathcal{L}_{\text{dist}} &= \operatorname*{mean}_{i,j}\ell_\delta\!\big(d_s(i,j),\,d_t(i,j)\big), \\
  u_z(i,j) &= \frac{z_j-z_i}{\max(\lVert z_j-z_i\rVert,\epsilon)}, \notag\\
  a_z(i,j,k) &= \big\langle u_z(i,j),u_z(i,k)\big\rangle, \notag\\
  \mathcal{L}_{\text{ang}} &= \operatorname*{mean}_{i,j,k}\ell_\delta\!\big(a_s(i,j,k),\,a_t(i,j,k)\big),
\end{align}
where $i,j,k$ index the $B$ class-token embeddings in a batch and the teacher relations are computed without gradients. The angle $a_z(i,j,k)$ is taken at anchor $i$ between the directions to $j$ and $k$; computed over all triples, this is the RKD triplet-angle term. Repeated-index cases are included for implementation simplicity, but they add no mismatch signal because the same degenerate relation is present on both sides. The constant $\epsilon=10^{-12}$ prevents division by zero, and $\mu_s,\mu_t$ are lower-bounded by $10^{-8}$ for the same reason. We use $\mathcal{L}_{\text{rkd}}=\mathcal{L}_{\text{dist}}+2\mathcal{L}_{\text{ang}}$, following the distance-to-angle ratio of RKD. Because the loss compares geometry within each representation space, it needs no projector and does not require equal feature dimensions.

\paragraph{Combined class-token loss.} The cosine loss is pointwise, aligning each tile's class token with its own teacher token, while RKD is relational, matching distances and angles across the batch without pinning any single token to a target. Because they impose different constraints, we use both on the class token,
\begin{equation}
  \mathcal{L}_{\text{cls}}^{(T)} = \mathcal{L}_{\text{cos}} + \gamma_T\,\mathcal{L}_{\text{rkd}},
\end{equation}
where $T$ indexes the teacher. The coefficient $\gamma_T$ is teacher-specific because the raw RKD scale differs across teacher-student pairs.

\paragraph{Patch-token loss.} A class-token loss backpropagates through the whole transformer, because the class token attends to the patch tokens, but it does not directly constrain the final patch outputs. We therefore add explicit patch supervision. For our patch-14 teachers, a $224\times224$ input produces a $16\times16$ grid, while the ViT-S/16 student produces a $14\times14$ grid. We resize the teacher grid to the student grid with PyTorch bicubic interpolation (\texttt{align\_corners=False}), denoted $\tilde{t}_p$, and use
\begin{equation}
  \mathcal{L}_{\text{patch}} = 1 - \cos\!\big(g(s_p),\, \tilde{t}_p\big),
\end{equation}
averaged over the batch and patch positions. This follows the patch term of the GPFM expert feature loss, with one deliberate change: GPFM adds a pointwise smooth-$L_1$ penalty on the patch features alongside the cosine term, and we keep the cosine term alone. We do not add a patch-level RKD term: after grid resizing, patch tokens have explicit spatial correspondences, so pointwise cosine is the direct supervision signal, and applying RKD across all patch positions would raise cost and dilute that spatial supervision. Unlike GPFM, which combines several teachers inside DINOv2 pretraining, we apply the loss to a single frozen teacher.

\paragraph{Full objective.} The main objective is
\begin{equation}
  \mathcal{L}^{(T)} = \mathcal{L}_{\text{cos}} + \gamma_T\,\mathcal{L}_{\text{rkd}} + \lambda_T\,\mathcal{L}_{\text{patch}},
\end{equation}
where $\lambda_T$ is also teacher-specific. The coefficients account for the different numerical ranges of the three loss terms. We use one standardized loss-contribution recipe for the four-teacher comparison: in the late training regime, RKD contributes approximately $25\%$ of the class-token loss and patch supervision contributes approximately $25\%$ of the total objective. We estimate the realized weighted-loss fractions as $\gamma_T\mathcal{L}_{\text{rkd}}/(\mathcal{L}_{\text{cos}}+\gamma_T\mathcal{L}_{\text{rkd}})$ and $\lambda_T\mathcal{L}_{\text{patch}}/\mathcal{L}^{(T)}$, averaged over the 40{,}000--50{,}000 step window. Table~\ref{tab:loss_weights} lists the coefficients and the realized late-window fractions for the four kaiko ViT-S/16 runs.

\begin{table}[ht]
\centering
\caption{Teacher-specific loss coefficients for the DistillPath-KS16 recipe. The class-token coefficient $\gamma_T$ scales RKD, and $\lambda_T$ scales the patch-token cosine loss. The last two columns report realized weighted-loss fractions averaged over steps 40{,}000--50{,}000.}
\label{tab:loss_weights}
\scriptsize
\setlength{\tabcolsep}{12pt}
\begin{tabular}{lcccc}
\toprule
Teacher & $\gamma_T$ & $\lambda_T$ & RKD / CLS & Patch / total \\
\midrule
H0-mini & 24 & 0.16 & 25.2\% & 28.3\% \\
UNI2-h & 32 & 0.31 & 24.4\% & 24.0\% \\
H-optimus-0 & 84 & 0.28 & 25.4\% & 24.6\% \\
Virchow2 & 56 & 0.30 & 24.5\% & 24.5\% \\
\bottomrule
\end{tabular}
\end{table}

\section{Experimental Setup}
\label{sec:setup}

\subsection{Distillation data}

We train on 6{,}000 TCGA H\&E whole-slide images from 32 cancer cohorts~\cite{tcga}; the number of slides per cohort follows the observed TCGA distribution and is not rebalanced. The training set is close in size to the 6{,}093-slide TCGA subset used by Phikon and H0-mini~\cite{phikon,h0mini}. At batch size 256 and 50{,}000 steps, each run uses 12.8 million accepted tile views. Grouping the scanner-reported microns-per-pixel values to the nearest nominal scale, the accepted tiles were $44.9\%$ at about $0.25$ mpp, $7.2\%$ at about $0.5$ mpp, $44.8\%$ at about $1.0$ mpp, $2.4\%$ at about $2.0$ mpp, and $0.7\%$ other or unknown.

\subsection{Online tile streaming}
\label{sec:online_streaming}

Following kaiko, we sample tiles directly from the slides during training rather than pre-extracting a fixed tile set~\cite{kaiko}, using the open-source wsistream library\footnote{\url{https://github.com/RamonKaspar/wsistream}}. To amortize whole-slide I/O, the sampler keeps a small pool of slides open and draws tiles across that pool before replacing slides. Slides are re-queued indefinitely, so training is step-based rather than epoch-based, and magnification, tissue, and color filters are applied online.

For each tile, tissue is detected on a low-resolution thumbnail with the CLAM detector and Otsu thresholding~\cite{clam,otsu}. We sample $256\times256$ tiles at $0.25$, $0.5$, $1.0$, and $2.0$ microns per pixel and keep only tiles with at least $40\%$ tissue. As in Midnight, we filter low-information tiles in HSV space and apply HED color augmentation~\cite{midnight}: a tile is kept only if at least $60\%$ of its pixels fall in the hue range $[90,180]$, saturation $[8,255]$, and value $[103,255]$. We set the HED strength to $\sigma=0.08$, resize each tile to $224\times224$, and pass the same augmented tile to both models, each with its own normalization statistics.

\subsection{Models and training}

The student is the 22M kaiko ViT-S/16, with output dimension 384, initialized from the public pathology-pretrained kaiko weights~\cite{kaiko}. The undistilled kaiko model is the baseline we compare against. The four teachers are H0-mini (86M, ViT-B/14)~\cite{h0mini}, Virchow2 (632M, ViT-H/14)~\cite{virchow2}, UNI2-h (681M, ViT-H/14)~\cite{uni2}, and H-optimus-0 (1.1B, ViT-g/14)~\cite{hoptimus0}, with output dimensions 768, 1280, 1536, and 1536.

We train each run for 50{,}000 steps with batch size 256 in bfloat16. We use AdamW~\cite{adamw} with learning rate $10^{-4}$, weight decay $0.05$, 500 warmup steps, cosine decay to $10^{-6}$, and gradient clipping at norm $3.0$. Each run took about 24--29 GPU-hours on one NVIDIA RTX 4090.

\subsection{Evaluation}

All teachers, baselines, and distilled checkpoints are evaluated with the same EVA, HEST, and PLISM protocols. EVA is a tile-level pathology benchmark suite with classification and segmentation tasks~\cite{eva}. We follow the official protocols on BACH~\cite{bach}, CRC~\cite{crc}, PCam~\cite{pcam}, MHIST~\cite{mhist}, BreakHis~\cite{breakhis}, Gleason~\cite{gleason}, CoNSeP~\cite{consep}, and MoNuSAC~\cite{monusac}. The CRC task uses NCT-CRC-HE-100K with CRC-VAL-HE-7K. Classification uses the class-token embedding and segmentation uses the last-block spatial feature map. Thus Virchow2 uses the same class-token classification interface as the other encoders, not model-specific pooled or concatenated embeddings. EVA classification tasks use balanced accuracy, and CoNSeP and MoNuSAC use MonaiDice. We report the mean over five probe runs. Following the EVA leaderboard, BACH is shown separately because its effective resolution after resizing is inconsistent with the other tasks, so the EVA mean covers the remaining seven tasks~\cite{evaleaderboard}. Across the five probe runs on each frozen encoder, the standard deviation of the EVA mean is at most $0.002$; this variance is from repeated downstream probing, as each distillation run was performed once.

HEST evaluates spatial-transcriptomics prediction across nine tasks~\cite{hest}. Class-token embeddings are reduced to 256 dimensions with PCA and used to fit ridge regression models for gene-expression prediction, and we report the mean Pearson correlation across tasks. PLISM measures how consistently a frozen encoder represents matched tissue under scanner and staining changes; we report the aggregate score on the reference 8{,}139-tile protocol~\cite{plism,h0mini}. Our analysis focuses on EVA and HEST, with PLISM as an additional robustness benchmark. Appendices~\ref{sec:supp_data}--\ref{sec:supp_results} provide full implementation details, hyperparameters, and per-checkpoint results for all eight runs.

\section{Results}
\label{sec:results}

\subsection{DistillPath-KS16 narrows the EVA gap to large encoders}

\begin{table}[ht]
\centering
\caption{EVA results for DistillPath-KS16 variants, distilled into the same 22M kaiko ViT-S/16. EVA$_{\text{mean}}$ averages the seven non-BACH tasks; BrHis, Gleas., and MoNu.\ denote BreakHis, Gleason, and MoNuSAC. Bold marks the best value in each column across all rows; underline marks the best DistillPath-KS16 variant in each column.}
\label{tab:eva}
\scriptsize
\setlength{\tabcolsep}{1.9pt}
\resizebox{\textwidth}{!}{%
\begin{tabular}{lccccccccc}
\toprule
& \multicolumn{6}{c}{EVA classification} & \multicolumn{2}{c}{EVA segmentation} & \\
\cmidrule(lr){2-7}\cmidrule(lr){8-9}
Model & BACH & PCam & CRC & MHIST & BrHis & Gleas. & CoNSeP & MoNu. & \textbf{EVA$_{\text{mean}}$} \\
\midrule
\multicolumn{10}{l}{\textit{Teachers, reference encoders, and baseline}} \\
Virchow2 (632M) & 0.879 & 0.939 & \textbf{0.966} & \textbf{0.861} & 0.821 & 0.778 & 0.640 & 0.667 & \textbf{0.810} \\
UNI2-h (681M) & \textbf{0.917} & \textbf{0.951} & \textbf{0.966} & 0.821 & \textbf{0.859} & 0.772 & 0.630 & 0.643 & 0.806 \\
H-optimus-0 (1.1B) & 0.756 & 0.942 & 0.956 & 0.843 & 0.806 & 0.752 & \textbf{0.642} & \textbf{0.681} & 0.803 \\
Midnight-12k (1.1B) & 0.900 & 0.929 & \textbf{0.966} & 0.799 & 0.816 & \textbf{0.799} & 0.624 & 0.658 & 0.799 \\
GPFM (303M) & 0.829 & 0.945 & 0.953 & 0.813 & 0.764 & 0.763 & 0.637 & 0.650 & 0.789 \\
H0-mini (86M) & 0.789 & 0.942 & 0.960 & 0.786 & 0.743 & 0.784 & 0.630 & 0.642 & 0.784 \\
kaiko ViT-S/16 baseline (22M) & 0.832 & 0.901 & 0.939 & 0.830 & 0.720 & 0.723 & 0.600 & 0.633 & 0.764 \\
\midrule
\multicolumn{10}{l}{\textit{DistillPath-KS16 variants (ours), all using the 22M kaiko ViT-S/16 architecture}} \\
DistillPath-KS16-Virchow2 & \underline{0.841} & 0.922 & \underline{0.957} & 0.811 & \underline{0.849} & \underline{0.774} & 0.618 & 0.633 & \underline{0.795} \\
DistillPath-KS16-HOpt0 & 0.742 & 0.921 & 0.943 & \underline{0.821} & 0.690 & 0.756 & 0.617 & \underline{0.634} & 0.769 \\
DistillPath-KS16-H0mini & 0.789 & \underline{0.927} & 0.951 & 0.820 & 0.714 & 0.743 & \underline{0.623} & 0.622 & 0.771 \\
DistillPath-KS16-UNI2h & 0.807 & \underline{0.927} & \underline{0.957} & 0.806 & 0.705 & 0.762 & 0.616 & 0.629 & 0.772 \\
\bottomrule
\end{tabular}
}
\end{table}

Table~\ref{tab:eva} reports EVA for the four teachers, Midnight-12k and GPFM as additional reference encoders, the kaiko baseline, and the four DistillPath-KS16 variants at the final 50{,}000-step checkpoint. Every variant improves the kaiko baseline of $0.764$ while keeping the 22M student architecture, although individual tasks can decrease. DistillPath-KS16-Virchow2 is the strongest variant at $0.795$, within $0.015$ points of the $0.810$ Virchow2 class-token reference at about $29\times$ fewer parameters, and it scores above H0-mini ($0.784$) and GPFM ($0.789$) on the aggregate EVA metric.

This aggregate advantage is concentrated in a few tasks rather than spread uniformly. Relative to kaiko, DistillPath-KS16-Virchow2 raises BreakHis from $0.720$ to $0.849$, above its own teacher's $0.821$ and the only task on which it beats its teacher, and raises Gleason from $0.723$ to $0.774$; it also improves PCam, CRC, and CoNSeP, while MHIST drops and MoNuSAC is essentially unchanged. The comparison to H0-mini and GPFM is therefore a seven-task summary rather than a per-task win: the advantage over both is driven primarily by BreakHis, with MHIST contributing against H0-mini and Gleason contributing against GPFM. Counting individual tasks, DistillPath-KS16-Virchow2 trails H0-mini on five and GPFM on four.

Teacher rank is not sufficient to predict transfer into this fixed student. H-optimus-0 is the largest teacher and scores $0.803$ EVA, yet DistillPath-KS16-HOpt0 has the lowest EVA mean among the four students at $0.769$. UNI2-h is the second strongest teacher at $0.806$, but gives a student close to H0-mini and below Virchow2. The HEST ordering is different again: H0-mini gives the strongest HEST student at $0.387$, while Virchow2 gives the weakest at $0.371$.

\begin{table}[ht]
\centering
\caption{HEST gene-expression prediction results. Values are Pearson correlations for the nine HEST tasks, followed by the mean. Bold marks the best value in each column across all rows; underline marks the best DistillPath-KS16 variant in each column.}
\label{tab:hest}
\scriptsize
\setlength{\tabcolsep}{2.0pt}
\resizebox{\textwidth}{!}{%
\begin{tabular}{lcccccccccc}
\toprule
Model & IDC & PRAD & PAAD & SKCM & COAD & READ & ccRCC & LUNG & LYMPH-IDC & \textbf{HEST$_{\text{mean}}$} \\
\midrule
\multicolumn{11}{l}{\textit{Teachers, reference encoders, and baseline}} \\
Virchow2 (632M) & 0.592 & 0.348 & 0.472 & 0.619 & 0.259 & 0.209 & \textbf{0.274} & 0.553 & 0.256 & 0.398 \\
UNI2-h (681M) & 0.590 & 0.357 & \textbf{0.500} & \textbf{0.659} & 0.301 & \textbf{0.223} & 0.264 & 0.558 & \textbf{0.272} & 0.414 \\
H-optimus-0 (1.1B) & \textbf{0.598} & \textbf{0.385} & 0.491 & 0.645 & \textbf{0.309} & 0.222 & 0.268 & \textbf{0.559} & 0.259 & \textbf{0.415} \\
Midnight-12k (1.1B) & 0.582 & 0.337 & 0.490 & 0.636 & 0.291 & 0.185 & 0.213 & 0.558 & 0.264 & 0.395 \\
GPFM (303M) & 0.566 & 0.342 & 0.460 & 0.589 & 0.248 & 0.165 & 0.259 & 0.547 & 0.237 & 0.379 \\
H0-mini (86M) & 0.586 & 0.368 & 0.492 & 0.601 & 0.249 & 0.186 & 0.267 & 0.548 & 0.263 & 0.396 \\
kaiko ViT-S/16 baseline (22M) & 0.533 & 0.348 & 0.441 & 0.545 & 0.206 & 0.133 & 0.210 & 0.503 & 0.225 & 0.349 \\
\midrule
\multicolumn{11}{l}{\textit{DistillPath-KS16 variants (ours), all using the 22M kaiko ViT-S/16 architecture}} \\
DistillPath-KS16-Virchow2 & 0.569 & 0.357 & 0.450 & 0.508 & 0.263 & 0.147 & 0.264 & 0.531 & 0.254 & 0.371 \\
DistillPath-KS16-HOpt0 & 0.554 & 0.349 & 0.458 & 0.554 & 0.261 & \underline{0.177} & 0.228 & \underline{0.546} & 0.256 & 0.376 \\
DistillPath-KS16-H0mini & \underline{0.573} & 0.361 & \underline{0.489} & 0.561 & \underline{0.277} & 0.165 & \underline{0.272} & 0.534 & 0.255 & \underline{0.387} \\
DistillPath-KS16-UNI2h & 0.562 & \underline{0.365} & 0.442 & \underline{0.571} & 0.243 & 0.151 & 0.248 & 0.534 & \underline{0.258} & 0.375 \\
\bottomrule
\end{tabular}
}
\end{table}

Table~\ref{tab:hest} shows a different pattern from the EVA results. All four DistillPath-KS16 variants improve over the kaiko HEST mean of $0.349$, but none reaches its corresponding teacher on the HEST mean or the standalone H0-mini reference mean of $0.396$. The best HEST student is DistillPath-KS16-H0mini at $0.387$, and its advantage comes from several tasks rather than one outlier: it is the best DistillPath-KS16 variant on IDC, PAAD, COAD, ccRCC, and the mean. DistillPath-KS16-Virchow2, despite being best on EVA, is the weakest HEST student at $0.371$ and is not the best variant on any individual HEST task.

\begin{table}[ht]
\centering
\caption{PLISM robustness results on the 8{,}139-tile protocol. PLISM score is the benchmark aggregate: the mean of all-pairs median cosine similarity and median top-10 retrieval accuracy for scanner-only, staining-only, and paired scanner-plus-staining changes. The remaining columns are diagnostic summaries: median cosine similarity and median top-5 retrieval accuracy. Bold marks the best value in each column across all rows; underline marks the best DistillPath-KS16 variant in each column.}
\label{tab:plism}
\scriptsize
\setlength{\tabcolsep}{3.0pt}
\resizebox{\textwidth}{!}{%
\begin{tabular}{lcccccc}
\toprule
Model & \shortstack{\textbf{PLISM}\\\textbf{score}} & \shortstack{All pairs\\cosine} & \shortstack{All pairs\\top-5} & \shortstack{Scanner\\top-5} & \shortstack{Stain\\top-5} & \shortstack{Scanner+stain\\top-5} \\
\midrule
\multicolumn{7}{l}{\textit{Teachers, reference encoders, and baseline}} \\
Virchow2 (632M) & 0.447 & 0.744 & 0.094 & 0.516 & 0.203 & 0.076 \\
UNI2-h (681M) & 0.333 & 0.592 & 0.033 & 0.421 & 0.123 & 0.023 \\
H-optimus-0 (1.1B) & 0.480 & 0.686 & 0.124 & 0.668 & \textbf{0.240} & 0.103 \\
Midnight-12k (1.1B) & 0.337 & 0.743 & 0.060 & 0.276 & 0.111 & 0.051 \\
GPFM (303M) & 0.264 & 0.594 & 0.009 & 0.253 & 0.049 & 0.006 \\
H0-mini (86M) & \textbf{0.540} & 0.800 & \textbf{0.135} & \textbf{0.798} & 0.224 & \textbf{0.111} \\
kaiko ViT-S/16 baseline (22M) & 0.307 & 0.756 & 0.020 & 0.248 & 0.067 & 0.014 \\
\midrule
\multicolumn{7}{l}{\textit{DistillPath-KS16 variants (ours), all using the 22M kaiko ViT-S/16 architecture}} \\
DistillPath-KS16-Virchow2 & 0.447 & 0.720 & 0.077 & 0.578 & 0.191 & 0.060 \\
DistillPath-KS16-HOpt0 & 0.480 & 0.656 & \underline{0.115} & \underline{0.738} & \underline{0.228} & \underline{0.093} \\
DistillPath-KS16-H0mini & \underline{0.495} & \underline{\textbf{0.816}} & 0.093 & 0.674 & 0.187 & 0.073 \\
DistillPath-KS16-UNI2h & 0.484 & 0.724 & 0.086 & 0.727 & 0.206 & 0.068 \\
\bottomrule
\end{tabular}
}
\end{table}

Table~\ref{tab:plism} shows that PLISM favors a different ordering. H0-mini remains the strongest reference encoder on the aggregate score, and the H0-mini-distilled student is the strongest of the four DistillPath-KS16 variants. Distillation improves the kaiko PLISM score from $0.307$ to $0.447$--$0.495$, but the strongest EVA model is not the strongest robustness model: the H-optimus-0-distilled student gives the best distilled top-5 retrieval under scanner and staining changes, while the H0-mini-distilled student gives the highest aggregate score and the highest median cosine similarity.

\subsection{Computational efficiency}

At inference, DistillPath-KS16 is a 22M ViT-S/16 encoder with 384-dimensional outputs. To quantify its computational cost relative to larger encoders, we measured encoder-forward throughput on TCGA tissue tiles using batch size 64 on two devices: one NVIDIA RTX 4090 with bfloat16, and a MacBook Pro with Apple M4 Pro using MPS in fp32. Table~\ref{tab:efficiency} reports mean throughput over timed batches after 30 warmup batches.

\begin{table}[ht]
\centering
\caption{Encoder-forward efficiency on TCGA tiles at batch size 64. The RTX 4090 benchmark uses CUDA/bfloat16 and 500 timed batches; the MacBook Pro benchmark uses Apple M4 Pro MPS/fp32 and 200 timed batches. Slowdown is relative to DistillPath-KS16 on the same device, so larger values are worse. CUDA peak reports peak allocated CUDA memory, while MPS alloc.\ reports allocated MPS memory recorded by the benchmark. FLOPs are estimates for linear, convolution, and ViT attention matmuls, excluding normalization, softmax, activation, residual, and preprocessing operations. Storage is fp32 feature storage per one million tile embeddings.}
\label{tab:efficiency}
\scriptsize
\setlength{\tabcolsep}{3pt}
\resizebox{\textwidth}{!}{%
\begin{tabular}{lccccccccccc}
\toprule
Model & Params & Dim & EVA$_{\text{mean}}$ & GFLOPs/tile & \multicolumn{2}{c}{RTX 4090} & \multicolumn{2}{c}{M4 Pro} & CUDA peak & MPS alloc. & Storage \\
 & & & & & tiles/s & slowdown & tiles/s & slowdown & (MiB) & (MiB) & (GiB/1M) \\
\midrule
DistillPath-KS16-Virchow2 & 21.7M & 384 & 0.795 & 9.2 & 7994 & $1.0\times$ & 396.1 & $1.0\times$ & 374 & 138 & 1.43 \\
H0-mini & 85.7M & 768 & 0.784 & 47.1 & 1807 & $4.4\times$ & 80.7 & $4.9\times$ & 1082 & 592 & 2.86 \\
GPFM & 303.2M & 1024 & 0.789 & 162.0 & 602 & $13.3\times$ & 26.6 & $14.9\times$ & 2388 & 1666 & 3.81 \\
Virchow2 & 631.2M & 1280 & 0.810 & 340.1 & 300 & $26.6\times$ & 12.8 & $30.9\times$ & 4551 & 2662 & 4.77 \\
H-optimus-0 & 1134.8M & 1536 & 0.803 & 608.3 & 181 & $44.2\times$ & 7.4 & $53.5\times$ & 7618 & 5146 & 5.72 \\
\bottomrule
\end{tabular}
}
\end{table}

On the RTX 4090, the best DistillPath-KS16 variant is faster than H0-mini, GPFM, Virchow2, and H-optimus-0 by $4.4\times$, $13.3\times$, $26.6\times$, and $44.2\times$, respectively. On the MacBook Pro, the same within-device comparisons are $4.9\times$, $14.9\times$, $30.9\times$, and $53.5\times$. It also uses $12.2\times$ less peak CUDA memory and $19.3\times$ less measured MPS allocated memory than Virchow2 in this batch-64 benchmark, and requires $3.3\times$ less fp32 storage for class-token features. The distilled encoder therefore narrows the EVA gap to larger encoders at lower forward-pass time, device memory, and feature-storage cost, while retaining the kaiko ViT-S/16 inference architecture.

\subsection{Teacher-dependent training dynamics}

\begin{figure}[ht]
  \centering
  \includegraphics[width=\textwidth]{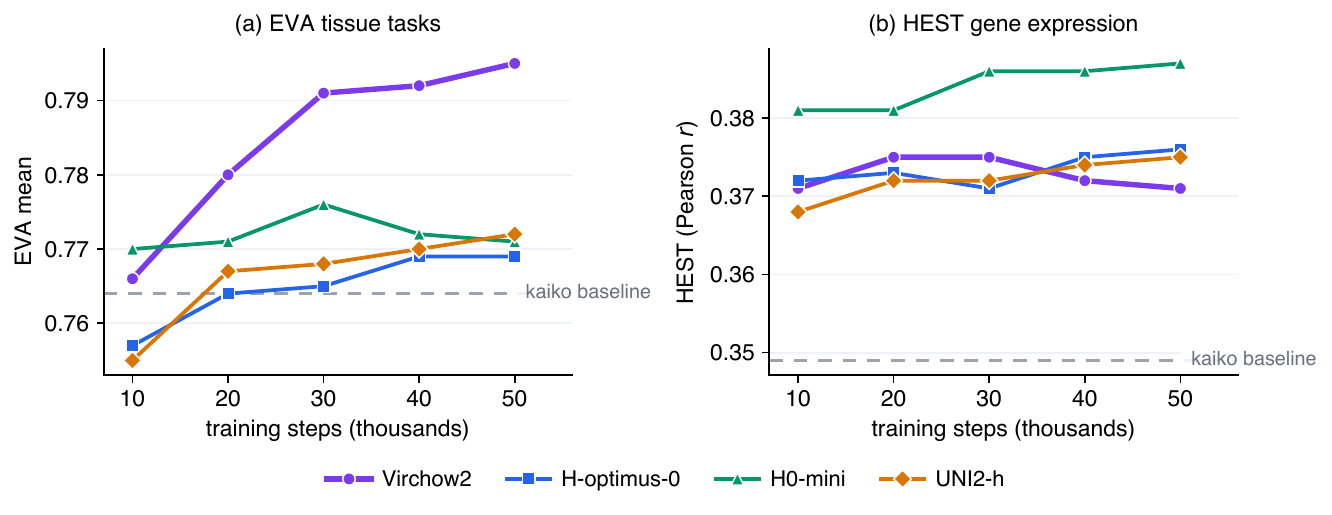}
  \caption{\textbf{Distillation trajectories} for the four teachers into kaiko ViT-S/16, across five checkpoints. (a) EVA mean over seven tasks; (b) mean HEST Pearson $r$ over nine tasks. The dashed line is the kaiko baseline. Virchow2 leads on EVA at later checkpoints, while H0-mini leads on HEST, so the best student depends on the target task.}
  \label{fig:trajectories}
\end{figure}

Figure~\ref{fig:trajectories} shows the EVA mean and HEST across the five saved checkpoints, and the four runs separate early. At 10{,}000 steps, H0-mini is strongest on both EVA and HEST, while Virchow2 is only slightly above the kaiko EVA baseline. By 20{,}000 steps, Virchow2 has become the strongest EVA student at $0.780$; H-optimus-0 is still only at the rounded kaiko baseline of $0.764$, so the first checkpoint at which all four runs exceed baseline is 30{,}000 steps. From 30{,}000 to 50{,}000 steps, each run varies by at most $0.005$ EVA, with Virchow2 remaining highest and ending at $0.795$.

The HEST panel differs from EVA. The H0-mini student is strongest at every checkpoint and improves from $0.381$ at 10{,}000 steps to $0.387$ at 50{,}000 steps, even though H0-mini is the lightest teacher and not the best on EVA. Virchow2 moves in the opposite direction late in training: it reaches $0.375$ HEST at 20{,}000--30{,}000 steps and then drops to $0.371$ by 50{,}000 steps while its EVA continues to improve. The HEST ranking of the students therefore does not match their EVA ranking, so no single DistillPath-KS16 variant is best on both axes, consistent with the two benchmarks emphasizing different targets: tissue classification and segmentation for EVA, and gene-expression prediction from the class token for HEST.

\subsection{Effect of student initialization}

To test whether DistillPath requires a pathology-pretrained student, we repeat the same backbone-token loss form and 50{,}000-step training setup with a 22M ViT-S/16 initialized from ImageNet-21k~\cite{imagenet21k} instead of kaiko (Table~\ref{tab:generality}). The ImageNet-21k baseline is weaker than kaiko on EVA and HEST, at $0.729$ versus $0.764$ EVA and $0.311$ versus $0.349$ HEST, but it has a higher PLISM score of $0.383$ versus $0.307$.

Within the same calibrated 25/25 setup, backbone-token distillation improves the ImageNet-initialized student for every teacher on all three aggregate metrics. EVA rises from $0.729$ to $0.754$--$0.768$, HEST from $0.311$ to $0.358$--$0.377$, and PLISM from $0.383$ to $0.490$--$0.561$. The teacher ordering differs from the kaiko-initialized runs: H-optimus-0 gives the best ImageNet-initialized EVA score at $0.768$, H0-mini the best HEST score at $0.377$, and UNI2-h the best PLISM score at $0.561$. The same loss can therefore transfer pathology signal into a generic pretrained ViT-S/16, but the preferred teacher depends on the student initialization and target benchmark.

The ImageNet-initialized runs also show the limit of this transfer within the calibrated setup. Even the best ImageNet-initialized EVA score is only around the undistilled kaiko baseline and remains well below DistillPath-KS16-Virchow2 at $0.795$. The gap is clearest on spatial EVA tasks: the distilled ImageNet-initialized students improve over their own baseline on CoNSeP and MoNuSAC, but remain below the kaiko baseline on MoNuSAC. PLISM behaves differently: UNI2-h distilled into the ImageNet-initialized student reaches $0.561$, above both the H0-mini reference and the kaiko-initialized DistillPath variants. Student initialization therefore changes not only the absolute transfer strength, but also which benchmark benefits most.

\begin{table}[H]
\centering
\caption{Student initialization comparison after the same 50{,}000-step distillation setup. IS16 denotes an ImageNet-21k ViT-S/16 student; DistillPath-IS16-X denotes the same ImageNet-initialized student distilled from teacher X. EVA$_{\text{mean}}$ averages the seven non-BACH EVA tasks as in Table~\ref{tab:eva}; BrHis, Gleas., and MoNu.\ denote BreakHis, Gleason, and MoNuSAC. HEST is the mean Pearson correlation across nine tasks, and PLISM is the aggregate robustness score.}
\label{tab:generality}
\scriptsize
\setlength{\tabcolsep}{1.1pt}
\resizebox{\textwidth}{!}{%
\begin{tabular}{lccccccccccc}
\toprule
& \multicolumn{6}{c}{EVA classification} & \multicolumn{2}{c}{EVA segmentation} & & & \\
\cmidrule(lr){2-7}\cmidrule(lr){8-9}
Model & BACH & PCam & CRC & MHIST & BrHis & Gleas. & CoNSeP & MoNu. & \textbf{EVA$_{\text{mean}}$} & \textbf{HEST} & \textbf{PLISM} \\
\midrule
kaiko ViT-S/16 baseline & 0.832 & 0.901 & 0.939 & 0.830 & 0.720 & 0.723 & 0.600 & 0.633 & 0.764 & 0.349 & 0.307 \\
ViT-S/16 IN21K baseline & 0.612 & 0.856 & 0.904 & 0.815 & 0.723 & 0.709 & 0.515 & 0.581 & 0.729 & 0.311 & 0.383 \\
\midrule
DistillPath-IS16-Virchow2 & 0.828 & 0.919 & 0.959 & 0.773 & 0.754 & 0.762 & 0.586 & 0.587 & 0.763 & 0.358 & 0.490 \\
DistillPath-IS16-HOpt0 & 0.705 & 0.917 & 0.951 & 0.796 & 0.765 & 0.736 & 0.599 & 0.611 & 0.768 & 0.363 & 0.526 \\
DistillPath-IS16-H0mini & 0.764 & 0.920 & 0.952 & 0.784 & 0.688 & 0.735 & 0.605 & 0.597 & 0.754 & 0.377 & 0.543 \\
DistillPath-IS16-UNI2h & 0.812 & 0.920 & 0.958 & 0.813 & 0.745 & 0.739 & 0.584 & 0.599 & 0.765 & 0.364 & 0.561 \\
\bottomrule
\end{tabular}
}
\end{table}

\section{Discussion}

DistillPath transfers signal from released pathology encoders into a compact student through backbone-token distillation alone. In the strongest case, Virchow2 distillation raises the 22M kaiko ViT-S/16 from $0.764$ to $0.795$ on the EVA mean while retaining the student's inference architecture and 384-dimensional feature size. As shown in Section~\ref{sec:results}, its aggregate EVA advantage over H0-mini and GPFM is task-concentrated rather than a uniform per-task win.

The four-teacher comparison also shows that transfer is not determined only by teacher size or teacher EVA. Virchow2 gives the best EVA student, H0-mini gives the best HEST and PLISM students, and the largest teacher, H-optimus-0, gives the lowest-EVA variant despite its strong teacher score. The divergent rankings across EVA, HEST, and PLISM indicate that these benchmarks reward different properties of the representation, and that the compact, robust H0-mini transfers gene-expression and robustness signal better than the larger teachers do. Teacher choice is therefore task-dependent, not a matter of using the largest or highest-scoring teacher.

The ImageNet-21k student experiments further show that teacher choice and student initialization interact. In the calibrated 25/25 setup, distillation improves a generic pretrained ViT-S/16 across EVA, HEST, and PLISM, but the best EVA score stays near the undistilled kaiko baseline. PLISM is the exception, with the UNI2-h ImageNet-initialized student exceeding the kaiko-initialized DistillPath variants, so student initialization shifts the tradeoff between tissue-task performance and robustness, not only the absolute transfer level.

The closest prior model, Virchow2G-Mini, is a similar 22M ViT-S obtained through DINOv2 distillation, but it uses the teacher's heads and one billion tiles~\cite{virchow2}. Our recipe is deliberately narrower: it trains from backbone features alone on 6{,}000 public slides, so it applies to released encoders that expose only final class and patch tokens. Public Virchow2G-Mini weights and matching benchmark outputs are not available, so we treat it as the closest conceptual comparison rather than an experimental baseline.

Several limitations remain. The main comparison uses one student architecture, the kaiko ViT-S/16. We include ImageNet-21k ViT-S experiments, but both students are pretrained ViT-S/16 models; we do not isolate how changes in student initialization, capacity, or architecture affect transfer, and we do not distill into a randomly initialized ViT-S/16. Because the main student has 384-dimensional outputs, we also cannot tell whether weaker transfer from the largest 1536-dimensional teachers reflects teacher-student mismatch, the training recipe, or a genuine capacity bottleneck; a ViT-Base student would be a natural next test. For a controlled comparison, we hold the evaluation interface fixed, which may understate teachers such as Virchow2, whose recommended feature extraction concatenates the class token with the mean of the patch tokens.

Several training and objective controls remain untested. We use one standardized loss-contribution balance for the main four-teacher comparison, but we do not present a full controlled ablation across all teachers, coefficients, schedules, and calibrated final settings. We compare with frozen kaiko but lack a no-teacher continued-training control and, unlike GPFM, do not combine feature matching with a self-supervised objective. Finally, all training slides come from TCGA, which is less diverse than the large collections used to train the teachers; results may depend on TCGA cohort composition, magnification distribution, and tissue filtering, and a more diverse or rebalanced training set might improve transfer. Multi-teacher distillation is another natural direction, but we isolate one teacher at a time throughout this work.

\section{Conclusion}

We presented DistillPath-KS16, a family of 22M kaiko ViT-S/16 encoders distilled from four released teachers (86M--1.1B parameters) using only frozen class and patch tokens and a combination of cosine and relational losses. Every distilled variant improves over the kaiko baseline on EVA, HEST, and PLISM, but the strongest teacher is task-dependent: Virchow2 gives the best EVA student, while H0-mini gives the best HEST and PLISM students. The best EVA variant, DistillPath-KS16-Virchow2, reaches $0.795$ on the EVA mean, within $0.015$ points of the Virchow2 class-token reference at about $29\times$ fewer parameters, and its aggregate EVA advantage over H0-mini and GPFM is driven primarily by BreakHis. The same loss also transfers pathology signal into a generic ImageNet-initialized ViT-S/16, though on EVA the kaiko-initialized students stay stronger. Relative to Virchow2, the EVA-best model is $26.6\times$ faster on an RTX 4090, $30.9\times$ faster on a MacBook Pro M4 Pro, and needs $3.3\times$ less fp32 feature storage. Overall, DistillPath is a practical, reproducible way to turn released pathology encoders into lower-cost students, using only frozen backbone tokens and public slides.


%
%
\bibliographystyle{splncs04}
\bibliography{main}

\clearpage
\appendix
\renewcommand{\theHsection}{appendix.\Alph{section}}
\renewcommand{\theHsubsection}{\theHsection.\arabic{subsection}}
\raggedbottom
\section{Training data composition}
\label{sec:supp_data}

\begin{figure}[htb]
\centering
\begin{tikzpicture}
\begin{axis}[
  xbar,
  width=0.47\textwidth,
  height=6.4cm,
  bar width=4pt,
  xmin=0,
  xmax=11.4,
  xtick={0,5,10},
  symbolic y coords={TGCT,COAD,HNSC,PRAD,STAD,BLCA,LUSC,SKCM,KIRC,THCA,LUAD,UCEC,SARC,LGG,GBM,BRCA},
  ytick=data,
  axis y line*=left,
  y axis line style={draw=none},
  axis x line*=bottom,
  xmajorgrids,
  grid style={wsigrid},
  tick style={draw=none},
  tick label style={font=\scriptsize},
  yticklabel style={text=black},
  xticklabel style={text=wsimuted},
  nodes near coords={\pgfmathprintnumber[fixed,precision=2,fixed zerofill]{\pgfplotspointmeta}},
  point meta=x,
  nodes near coords style={font=\tiny,text=wsimuted,anchor=west,xshift=1pt},
  enlarge y limits=0.035,
  clip=false
]
\addplot[fill=wsiviolet,fill opacity=1,draw=none] coordinates {(3.25,TGCT) (3.47,COAD) (3.63,HNSC) (3.85,PRAD) (3.88,STAD) (3.93,BLCA) (4.17,LUSC) (4.20,SKCM) (4.37,KIRC) (4.40,THCA) (4.65,LUAD) (4.72,UCEC) (5.15,SARC) (6.80,LGG) (7.22,GBM) (9.77,BRCA)};
\end{axis}
\end{tikzpicture}\hfill
\begin{tikzpicture}
\begin{axis}[
  xbar,
  width=0.47\textwidth,
  height=6.4cm,
  bar width=4pt,
  xmin=0,
  xmax=11.4,
  xtick={0,5,10},
  symbolic y coords={CHOL,DLBC,UVM,UCS,MESO,OV,KICH,THYM,ESCA,READ,PCPG,ACC,PAAD,CESC,KIRP,LIHC},
  ytick=data,
  axis y line*=left,
  y axis line style={draw=none},
  axis x line*=bottom,
  xmajorgrids,
  grid style={wsigrid},
  tick style={draw=none},
  tick label style={font=\scriptsize},
  yticklabel style={text=black},
  xticklabel style={text=wsimuted},
  nodes near coords={\pgfmathprintnumber[fixed,precision=2,fixed zerofill]{\pgfplotspointmeta}},
  point meta=x,
  nodes near coords style={font=\tiny,text=wsimuted,anchor=west,xshift=1pt},
  enlarge y limits=0.035,
  clip=false
]
\addplot[fill=wsiviolet,fill opacity=1,draw=none] coordinates {(0.37,CHOL) (0.43,DLBC) (0.72,UVM) (0.73,UCS) (0.80,MESO) (0.92,OV) (0.98,KICH) (1.32,THYM) (1.32,ESCA) (1.40,READ) (1.72,PCPG) (1.78,ACC) (2.02,PAAD) (2.23,CESC) (2.52,KIRP) (3.20,LIHC)};
\end{axis}
\end{tikzpicture}
\caption{TCGA cohort distribution in the 6{,}000-slide distillation set, ordered by share. The six slides without cohort labels are omitted from the bars; percentages use all 6{,}000 slides as the denominator.}
\label{fig:supp_cohorts}
\end{figure}
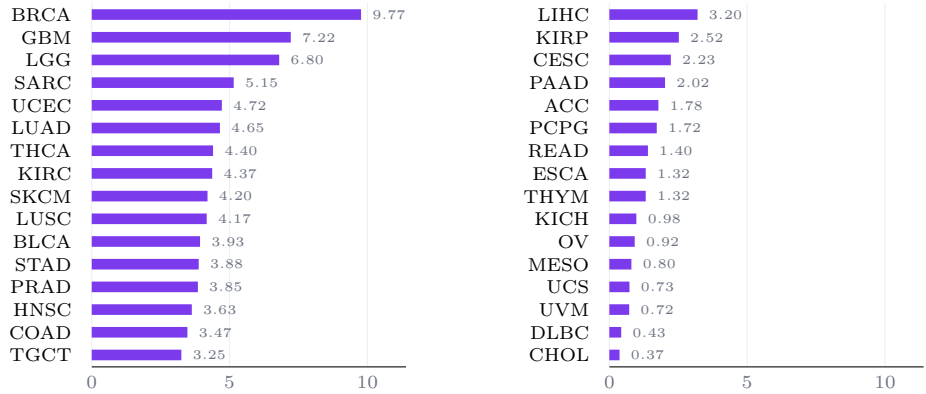

\begin{figure}[htb]
  \centering
  \includegraphics[width=0.78\textwidth]{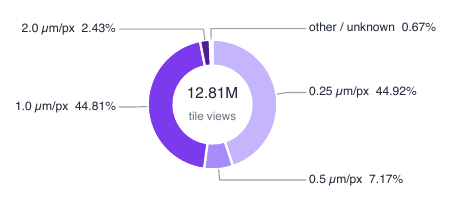}
  \caption{Share of accepted tile views per magnification among the 12.81 million views logged at step 50{,}000 of the DistillPath-KS16-Virchow2 run. Scanner-reported microns-per-pixel values are grouped into nominal bins spanning $[0.10,0.35)$, $[0.35,0.70)$, $[0.70,1.35)$, and $[1.35,2.35)$, respectively; values outside these intervals and missing metadata form the \emph{other / unknown} bin.}
  \label{fig:supp_magnification}
\end{figure}

The distillation set contains 6{,}000 TCGA H\&E whole-slide images spanning all 32 TCGA cohorts; cohort frequencies follow the observed TCGA distribution without rebalancing. Six slides lack cohort labels and are omitted from Figure~\ref{fig:supp_cohorts}. Because the sampler re-queues slides indefinitely instead of iterating over a fixed tile set, training is organized by steps rather than epochs (Section~\ref{sec:online_streaming}). Figure~\ref{fig:supp_magnification} reports the magnification distribution logged by the 50{,}000-step DistillPath-KS16-Virchow2 run.

\FloatBarrier

\section{Online tile streaming with wsistream}
\label{sec:supp_wsistream}

DistillPath uses \texttt{wsistream} 0.1.5\footnote{Source code: \url{https://github.com/RamonKaspar/wsistream/tree/v0.1.5}}, an MIT-licensed library we developed for online tile streaming from whole-slide images. Rather than extracting and storing a fixed tile corpus before training, \texttt{wsistream} constructs each training input on demand. Coordinate selection, target resolution, tissue detection, filtering, and augmentation therefore remain part of the experiment configuration.

The library organizes the WSI-to-tensor path as a sequence of configurable components. A slide backend exposes the image pyramid and its metadata, a tissue detector identifies eligible regions from a low-resolution thumbnail, and a sampler selects a location and target resolution. The corresponding image region is then read, filtered by pixel content, transformed, and returned with its metadata through a PyTorch iterable dataset. Separate component interfaces allow the slide reader, tissue detector, sampler, filter, and transforms to be changed without modifying the training loop.

Opening a slide and constructing its tissue mask incur fixed costs, while reading many consecutive tiles creates long single-slide runs. Each data-loading worker therefore maintains a bounded pool of open slides and visits them in round-robin order, balancing cost amortization with slide interleaving. The pool size bounds open slide handles, the per-visit budget controls how frequently the active slide changes, and the per-slide budget controls when a slide is closed and replaced. Table~\ref{tab:supp_stream} specifies the complete \texttt{wsistream} configuration used for DistillPath. The library is distributed through PyPI\footnote{\url{https://pypi.org/project/wsistream/0.1.5/}}, with API and usage documentation on its documentation website.\footnote{\url{https://ramonkaspar.github.io/wsistream/}}

\begin{table}[H]
\centering
\caption{\texttt{wsistream} 0.1.5 configuration used for DistillPath.}
\label{tab:supp_stream}
\footnotesize
\begin{tabular}{@{}l >{\raggedright\arraybackslash}p{7.1cm}@{}}
\toprule
Setting & Value \\
\midrule
Slide reader & TiffSlide \\
Tissue mask & CLAM detector with Otsu thresholding on a thumbnail bounded to $2048 \times 2048$ px \\
CLAM contour filtering & median-blur kernel $7$, closing kernel $4$, $a_t=16$, $a_h=4$, at most 8 holes per contour, reference patch size 512 \\
Magnification sampling & $0.25$, $0.5$, $1.0$, or $2.0$ microns per pixel, selected uniformly \\
Pyramid-level selection & nearest available level to the target scale; level 0 when slide-scale metadata are unavailable \\
Spatial sampling & random coordinates sampled with replacement; at least $40\%$ tissue within the candidate tile \\
Extracted tile & $256 \times 256$ px at the selected pyramid level \\
HSV tile filter & at least $60\%$ of pixels within hue $[90,180]$, saturation $[8,255]$, and value $[103,255]$ \\
HED augmentation & independently per channel and tile, $\alpha \sim \mathcal{U}(0.92,1.08)$ and $\beta \sim \mathcal{U}(-0.08,0.08)$ \\
Resize and tensor conversion & $224 \times 224$ with bilinear interpolation; float32 RGB scaled to $[0,1]$ \\
Teacher/student view & the same HED-augmented tile is passed to both models \\
Slide order & randomized and cycled indefinitely \\
Slide pool & five open slides per worker; 16 data-loading workers \\
Per-slide budget & 100 patch-read attempts before rotating the slide out of the pool \\
Per-slide visit & 8 patch-read attempts before moving to the next slide in the pool \\
\bottomrule
\end{tabular}
\end{table}

\section{Distillation implementation}
\label{sec:supp_hparams}

The eight runs pair four frozen teachers with two ViT-S/16 student initializations. The teacher, student initialization, and two loss coefficients vary by run; the objective, optimization schedule, and online data pipeline are otherwise shared. Tables~\ref{tab:supp_models}--\ref{tab:supp_optim} specify the shared and run-specific settings. The released repository\footnote{\url{https://github.com/RamonKaspar/DistillPath}} contains the training code and all eight experiment configurations.

\subsection{Encoder configurations}

\begin{table}[H]
\centering
\caption{Encoder configurations. All encoders use $224 \times 224$ inputs during distillation and evaluation. The same augmented tile is passed to teacher and student, and each model wrapper applies its own normalization statistics. Register tokens are excluded from the distillation loss.}
\label{tab:supp_models}
\footnotesize
\setlength{\tabcolsep}{4pt}
\begin{tabular}{llccll}
\toprule
Role & Model & Dim & Patch & Registers & Normalization preset \\
\midrule
Student & kaiko ViT-S/16 & 384 & 16 & 0 & kaiko \\
Student & ViT-S/16 IN21K & 384 & 16 & 0 & ImageNet \\
\midrule
Teacher & H0-mini & 768 & 14 & 4 & H-optimus \\
Teacher & Virchow2 & 1280 & 14 & 4 & ImageNet \\
Teacher & UNI2-h & 1536 & 14 & 8 & ImageNet \\
Teacher & H-optimus-0 & 1536 & 14 & 4 & H-optimus \\
\bottomrule
\end{tabular}
\end{table}

Kaiko uses $\mu=\sigma=(0.5,0.5,0.5)$. ImageNet uses $\mu=(0.485,0.456,0.406)$ and $\sigma=(0.229,0.224,0.225)$. H-optimus uses $\mu=(0.707223,0.578729,0.703617)$ and $\sigma=(0.211883,0.230117,0.177517)$.

\subsection{Distillation objective and loss coefficients}

\begin{table}[H]
\centering
\caption{Distillation objective and projector. The projector maps student tokens into the teacher dimension $d_t$ for the class- and patch-token cosine losses and is discarded after training. RKD is computed within the original student and teacher spaces and does not use the projector.}
\label{tab:supp_objective}
\footnotesize
\begin{tabular}{@{}l >{\raggedright\arraybackslash}p{7.1cm}@{}}
\toprule
Setting & Value \\
\midrule
Class-token loss & cosine $+$ RKD \\
Class-token cosine coefficient & $1.0$ \\
RKD coefficient $\gamma_T$ & run-specific (Table~\ref{tab:supp_coefficients}) \\
RKD distance term weight & $1.0$ \\
RKD angle term weight & $2.0$ \\
RKD penalty & Huber (smooth $L_1$), $\delta = 1$ \\
RKD distance normalization & mean off-diagonal, clamped at $10^{-8}$ \\
RKD direction $\epsilon$ & $10^{-12}$ \\
Patch-token loss & cosine \\
Patch coefficient $\lambda_T$ & run-specific (Table~\ref{tab:supp_coefficients}) \\
Patch-token grid resampling & bicubic, \texttt{align\_corners=False} \\
Loss dtype & terms computed in fp32 \\
\midrule
Projector & DINO-style head \\
\quad MLP & $384 \to 2048 \to$ GELU $\to 2048 \to$ GELU $\to 256$ \\
\quad Bottleneck & $\ell_2$ normalization at 256-d \\
\quad Output layer & weight-normalized linear $256 \to d_t$, no bias \\
\quad MLP linear layers & weights initialized from a truncated normal with standard deviation $0.02$; biases initialized to $0$ \\
\bottomrule
\end{tabular}
\end{table}

\begin{table}[H]
\centering
\caption{Loss coefficients used in each teacher--student run. The class-token coefficient $\gamma_T$ scales RKD, and $\lambda_T$ scales the patch-token cosine loss.}
\label{tab:supp_coefficients}
\footnotesize
\setlength{\tabcolsep}{8pt}
\begin{tabular}{llcc}
\toprule
Student initialization & Teacher & $\gamma_T$ & $\lambda_T$ \\
\midrule
kaiko ViT-S/16 & H0-mini & 24 & 0.16 \\
 & Virchow2 & 56 & 0.30 \\
 & UNI2-h & 32 & 0.31 \\
 & H-optimus-0 & 84 & 0.28 \\
\midrule
ViT-S/16 IN21K & H0-mini & 24 & 0.16 \\
 & Virchow2 & 28 & 0.39 \\
 & UNI2-h & 32 & 0.31 \\
 & H-optimus-0 & 65 & 0.28 \\
\bottomrule
\end{tabular}
\end{table}

\subsection{Optimization and schedule}

\begin{table}[H]
\centering
\caption{Optimization settings and schedule.}
\label{tab:supp_optim}
\footnotesize
\begin{tabular}{@{}l >{\raggedright\arraybackslash}p{7.1cm}@{}}
\toprule
Setting & Value \\
\midrule
Optimizer & AdamW \\
$\beta_1,\ \beta_2$ & $0.9,\ 0.999$ (PyTorch default) \\
$\epsilon$ & $10^{-8}$ (PyTorch default) \\
Peak learning rate & $10^{-4}$ \\
Final learning rate & $10^{-6}$ \\
Weight decay & $0.05$ \\
Weight-decay exclusions & parameters with $\text{ndim}\le 1$ (including biases and normalization gains), and \texttt{cls\_token}, \texttt{pos\_embed}, \texttt{reg\_token}, \texttt{dist\_token} \\
Warmup & 500 steps, linear from $0.01\times$ peak \\
Decay & cosine over the remaining 49{,}500 steps \\
Gradient clipping & global norm $3.0$, over student and projector \\
Training steps & 50{,}000 \\
Batch size & 256 \\
Gradient accumulation & 1 \\
Precision & bfloat16 autocast (fp32 master weights) \\
\bottomrule
\end{tabular}
\end{table}

\section{Full per-checkpoint results}
\label{sec:supp_results}

We report every saved checkpoint (10{,}000--50{,}000 steps) for all eight DistillPath runs, together with the reference encoders and baselines, on every task-level and aggregate metric used in the main paper. All values use the same evaluation protocols as the main paper. EVA and HEST report per-task scores followed by the mean, and PLISM reports the aggregate score followed by the diagnostic columns from the main-paper PLISM table. Model names are abbreviated: \mbox{KS16-$X$} denotes \mbox{DistillPath-KS16-$X$} (kaiko ViT-S/16 student) and \mbox{IS16-$X$} denotes \mbox{DistillPath-IS16-$X$} (ImageNet-21k ViT-S/16 student).

\clearpage

\begingroup
\fontsize{6}{7}\selectfont
\setlength{\tabcolsep}{1.5pt}
\begin{longtable}{llccccccccc}
\caption{Full EVA results across all checkpoints. EVA$_{\text{mean}}$ averages the seven non-BACH tasks.}\label{tab:supp_eva}\\
\toprule
Model & Step & BACH & PCam & CRC & MHIST & BrHis & Gleas. & CoNSeP & MoNu. & \textbf{EVA$_{\text{mean}}$} \\
\midrule
\endfirsthead
\multicolumn{11}{l}{\emph{\tablename\ \thetable\ (continued)}} \\
\toprule
Model & Step & BACH & PCam & CRC & MHIST & BrHis & Gleas. & CoNSeP & MoNu. & \textbf{EVA$_{\text{mean}}$} \\
\midrule
\endhead
\bottomrule
\endlastfoot
\multicolumn{11}{l}{\textit{Reference encoders and baselines}} \\
Virchow2 (632M) & -- & 0.879 & 0.939 & 0.966 & 0.861 & 0.821 & 0.778 & 0.640 & 0.667 & 0.810 \\
UNI2-h (681M) & -- & 0.917 & 0.951 & 0.966 & 0.821 & 0.859 & 0.772 & 0.630 & 0.643 & 0.806 \\
H-optimus-0 (1.1B) & -- & 0.756 & 0.942 & 0.956 & 0.843 & 0.806 & 0.752 & 0.642 & 0.681 & 0.803 \\
Midnight-12k (1.1B) & -- & 0.900 & 0.929 & 0.966 & 0.799 & 0.816 & 0.799 & 0.624 & 0.658 & 0.799 \\
GPFM (303M) & -- & 0.829 & 0.945 & 0.953 & 0.813 & 0.764 & 0.763 & 0.637 & 0.650 & 0.789 \\
H0-mini (86M) & -- & 0.789 & 0.942 & 0.960 & 0.786 & 0.743 & 0.784 & 0.630 & 0.642 & 0.784 \\
kaiko ViT-S/16 (22M) & -- & 0.832 & 0.901 & 0.939 & 0.830 & 0.720 & 0.723 & 0.600 & 0.633 & 0.764 \\
ViT-S/16 IN21K & -- & 0.612 & 0.856 & 0.904 & 0.815 & 0.723 & 0.709 & 0.515 & 0.581 & 0.729 \\
\midrule
\multicolumn{11}{l}{\textit{DistillPath-KS16 (kaiko ViT-S/16 student)}} \\
KS16-Virchow2 & 10k & 0.794 & 0.917 & 0.950 & 0.776 & 0.719 & 0.773 & 0.611 & 0.613 & 0.766 \\
 & 20k & 0.859 & 0.914 & 0.952 & 0.791 & 0.806 & 0.765 & 0.610 & 0.621 & 0.780 \\
 & 30k & 0.833 & 0.919 & 0.950 & 0.786 & 0.873 & 0.774 & 0.615 & 0.620 & 0.791 \\
 & 40k & 0.844 & 0.920 & 0.956 & 0.812 & 0.827 & 0.777 & 0.618 & 0.635 & 0.792 \\
 & 50k & 0.841 & 0.922 & 0.957 & 0.811 & 0.849 & 0.774 & 0.618 & 0.633 & 0.795 \\
\addlinespace
KS16-HOpt0 & 10k & 0.675 & 0.914 & 0.937 & 0.785 & 0.695 & 0.740 & 0.604 & 0.626 & 0.757 \\
 & 20k & 0.702 & 0.916 & 0.942 & 0.800 & 0.694 & 0.757 & 0.612 & 0.624 & 0.764 \\
 & 30k & 0.781 & 0.919 & 0.938 & 0.812 & 0.679 & 0.762 & 0.615 & 0.632 & 0.765 \\
 & 40k & 0.768 & 0.920 & 0.942 & 0.822 & 0.686 & 0.759 & 0.616 & 0.634 & 0.769 \\
 & 50k & 0.742 & 0.921 & 0.943 & 0.821 & 0.690 & 0.756 & 0.617 & 0.634 & 0.769 \\
\addlinespace
KS16-H0mini & 10k & 0.759 & 0.925 & 0.945 & 0.813 & 0.743 & 0.731 & 0.620 & 0.611 & 0.770 \\
 & 20k & 0.754 & 0.923 & 0.954 & 0.817 & 0.733 & 0.737 & 0.616 & 0.616 & 0.771 \\
 & 30k & 0.757 & 0.925 & 0.949 & 0.828 & 0.753 & 0.741 & 0.622 & 0.613 & 0.776 \\
 & 40k & 0.777 & 0.927 & 0.950 & 0.815 & 0.717 & 0.743 & 0.623 & 0.626 & 0.772 \\
 & 50k & 0.789 & 0.927 & 0.951 & 0.820 & 0.714 & 0.743 & 0.623 & 0.622 & 0.771 \\
\addlinespace
KS16-UNI2h & 10k & 0.770 & 0.910 & 0.948 & 0.770 & 0.695 & 0.734 & 0.606 & 0.619 & 0.755 \\
 & 20k & 0.820 & 0.920 & 0.952 & 0.799 & 0.712 & 0.759 & 0.613 & 0.617 & 0.767 \\
 & 30k & 0.825 & 0.923 & 0.956 & 0.794 & 0.697 & 0.769 & 0.608 & 0.628 & 0.768 \\
 & 40k & 0.805 & 0.927 & 0.957 & 0.809 & 0.695 & 0.757 & 0.615 & 0.633 & 0.770 \\
 & 50k & 0.807 & 0.927 & 0.957 & 0.806 & 0.705 & 0.762 & 0.616 & 0.629 & 0.772 \\
\midrule
\multicolumn{11}{l}{\textit{DistillPath-IS16 (ImageNet-21k ViT-S/16 student)}} \\
IS16-Virchow2 & 10k & 0.806 & 0.903 & 0.952 & 0.767 & 0.834 & 0.755 & 0.575 & 0.583 & 0.767 \\
 & 20k & 0.842 & 0.909 & 0.955 & 0.776 & 0.773 & 0.764 & 0.580 & 0.592 & 0.764 \\
 & 30k & 0.833 & 0.909 & 0.955 & 0.768 & 0.752 & 0.767 & 0.580 & 0.593 & 0.761 \\
 & 40k & 0.826 & 0.918 & 0.954 & 0.771 & 0.758 & 0.759 & 0.587 & 0.591 & 0.763 \\
 & 50k & 0.828 & 0.919 & 0.959 & 0.773 & 0.754 & 0.762 & 0.586 & 0.587 & 0.763 \\
\addlinespace
IS16-HOpt0 & 10k & 0.713 & 0.899 & 0.942 & 0.798 & 0.774 & 0.722 & 0.574 & 0.596 & 0.758 \\
 & 20k & 0.725 & 0.906 & 0.947 & 0.808 & 0.769 & 0.725 & 0.587 & 0.605 & 0.764 \\
 & 30k & 0.720 & 0.910 & 0.948 & 0.805 & 0.773 & 0.731 & 0.586 & 0.604 & 0.765 \\
 & 40k & 0.717 & 0.917 & 0.950 & 0.800 & 0.757 & 0.735 & 0.597 & 0.608 & 0.766 \\
 & 50k & 0.705 & 0.917 & 0.951 & 0.796 & 0.765 & 0.736 & 0.599 & 0.611 & 0.768 \\
\addlinespace
IS16-H0mini & 10k & 0.716 & 0.913 & 0.949 & 0.825 & 0.731 & 0.711 & 0.595 & 0.589 & 0.759 \\
 & 20k & 0.743 & 0.912 & 0.949 & 0.800 & 0.698 & 0.721 & 0.602 & 0.600 & 0.755 \\
 & 30k & 0.744 & 0.916 & 0.952 & 0.789 & 0.698 & 0.727 & 0.605 & 0.598 & 0.755 \\
 & 40k & 0.760 & 0.920 & 0.952 & 0.787 & 0.693 & 0.736 & 0.606 & 0.597 & 0.756 \\
 & 50k & 0.764 & 0.920 & 0.952 & 0.784 & 0.688 & 0.735 & 0.605 & 0.597 & 0.754 \\
\addlinespace
IS16-UNI2h & 10k & 0.797 & 0.907 & 0.942 & 0.789 & 0.720 & 0.733 & 0.573 & 0.587 & 0.750 \\
 & 20k & 0.792 & 0.917 & 0.957 & 0.805 & 0.719 & 0.726 & 0.581 & 0.597 & 0.757 \\
 & 30k & 0.801 & 0.919 & 0.953 & 0.813 & 0.723 & 0.738 & 0.584 & 0.602 & 0.761 \\
 & 40k & 0.805 & 0.921 & 0.957 & 0.810 & 0.741 & 0.733 & 0.582 & 0.600 & 0.763 \\
 & 50k & 0.812 & 0.920 & 0.958 & 0.813 & 0.745 & 0.739 & 0.584 & 0.599 & 0.765 \\
\end{longtable}
\endgroup

\begingroup
\fontsize{6}{7}\selectfont
\setlength{\tabcolsep}{0.5pt}
\begin{longtable}{llcccccccccc}
\caption{Full HEST gene-expression results (Pearson $r$) across all checkpoints.}\label{tab:supp_hest}\\
\toprule
Model & Step & IDC & PRAD & PAAD & SKCM & COAD & READ & ccRCC & LUNG & LYMPH-IDC & \textbf{HEST$_{\text{mean}}$} \\
\midrule
\endfirsthead
\multicolumn{12}{l}{\emph{\tablename\ \thetable\ (continued)}} \\
\toprule
Model & Step & IDC & PRAD & PAAD & SKCM & COAD & READ & ccRCC & LUNG & LYMPH-IDC & \textbf{HEST$_{\text{mean}}$} \\
\midrule
\endhead
\bottomrule
\endlastfoot
\multicolumn{12}{l}{\textit{Reference encoders and baselines}} \\
Virchow2 (632M) & -- & 0.592 & 0.348 & 0.472 & 0.619 & 0.259 & 0.209 & 0.274 & 0.553 & 0.256 & 0.398 \\
UNI2-h (681M) & -- & 0.590 & 0.357 & 0.500 & 0.659 & 0.301 & 0.223 & 0.264 & 0.558 & 0.272 & 0.414 \\
H-optimus-0 (1.1B) & -- & 0.598 & 0.385 & 0.491 & 0.645 & 0.309 & 0.222 & 0.268 & 0.559 & 0.259 & 0.415 \\
Midnight-12k (1.1B) & -- & 0.582 & 0.337 & 0.490 & 0.636 & 0.291 & 0.185 & 0.213 & 0.558 & 0.264 & 0.395 \\
GPFM (303M) & -- & 0.566 & 0.342 & 0.460 & 0.589 & 0.248 & 0.165 & 0.259 & 0.547 & 0.237 & 0.379 \\
H0-mini (86M) & -- & 0.586 & 0.368 & 0.492 & 0.601 & 0.249 & 0.186 & 0.267 & 0.548 & 0.263 & 0.396 \\
kaiko ViT-S/16 (22M) & -- & 0.533 & 0.348 & 0.441 & 0.545 & 0.206 & 0.133 & 0.210 & 0.503 & 0.225 & 0.349 \\
ViT-S/16 IN21K & -- & 0.467 & 0.276 & 0.381 & 0.461 & 0.223 & 0.091 & 0.156 & 0.502 & 0.239 & 0.311 \\
\midrule
\multicolumn{12}{l}{\textit{DistillPath-KS16 (kaiko ViT-S/16 student)}} \\
KS16-Virchow2 & 10k & 0.550 & 0.314 & 0.473 & 0.547 & 0.263 & 0.154 & 0.242 & 0.545 & 0.249 & 0.371 \\
 & 20k & 0.553 & 0.342 & 0.458 & 0.558 & 0.260 & 0.160 & 0.254 & 0.537 & 0.251 & 0.375 \\
 & 30k & 0.563 & 0.352 & 0.449 & 0.537 & 0.266 & 0.155 & 0.256 & 0.540 & 0.252 & 0.375 \\
 & 40k & 0.568 & 0.353 & 0.455 & 0.516 & 0.262 & 0.142 & 0.268 & 0.533 & 0.252 & 0.372 \\
 & 50k & 0.569 & 0.357 & 0.450 & 0.508 & 0.263 & 0.147 & 0.264 & 0.531 & 0.254 & 0.371 \\
\addlinespace
KS16-HOpt0 & 10k & 0.537 & 0.340 & 0.455 & 0.572 & 0.254 & 0.158 & 0.242 & 0.537 & 0.249 & 0.372 \\
 & 20k & 0.539 & 0.342 & 0.447 & 0.557 & 0.273 & 0.161 & 0.230 & 0.555 & 0.254 & 0.373 \\
 & 30k & 0.552 & 0.337 & 0.448 & 0.544 & 0.253 & 0.170 & 0.236 & 0.547 & 0.257 & 0.371 \\
 & 40k & 0.553 & 0.345 & 0.456 & 0.547 & 0.266 & 0.174 & 0.227 & 0.548 & 0.255 & 0.375 \\
 & 50k & 0.554 & 0.349 & 0.458 & 0.554 & 0.261 & 0.177 & 0.228 & 0.546 & 0.256 & 0.376 \\
\addlinespace
KS16-H0mini & 10k & 0.556 & 0.366 & 0.472 & 0.548 & 0.276 & 0.149 & 0.265 & 0.541 & 0.255 & 0.381 \\
 & 20k & 0.564 & 0.349 & 0.478 & 0.560 & 0.266 & 0.162 & 0.267 & 0.533 & 0.253 & 0.381 \\
 & 30k & 0.569 & 0.367 & 0.488 & 0.553 & 0.274 & 0.163 & 0.271 & 0.536 & 0.253 & 0.386 \\
 & 40k & 0.572 & 0.354 & 0.488 & 0.561 & 0.274 & 0.161 & 0.273 & 0.537 & 0.253 & 0.386 \\
 & 50k & 0.573 & 0.361 & 0.489 & 0.561 & 0.277 & 0.165 & 0.272 & 0.534 & 0.255 & 0.387 \\
\addlinespace
KS16-UNI2h & 10k & 0.535 & 0.344 & 0.452 & 0.576 & 0.238 & 0.159 & 0.227 & 0.530 & 0.253 & 0.368 \\
 & 20k & 0.552 & 0.352 & 0.447 & 0.571 & 0.256 & 0.133 & 0.240 & 0.541 & 0.252 & 0.372 \\
 & 30k & 0.554 & 0.368 & 0.436 & 0.564 & 0.237 & 0.161 & 0.239 & 0.538 & 0.248 & 0.372 \\
 & 40k & 0.561 & 0.362 & 0.437 & 0.579 & 0.236 & 0.152 & 0.247 & 0.535 & 0.259 & 0.374 \\
 & 50k & 0.562 & 0.365 & 0.442 & 0.571 & 0.243 & 0.151 & 0.248 & 0.534 & 0.258 & 0.375 \\
\midrule
\multicolumn{12}{l}{\textit{DistillPath-IS16 (ImageNet-21k ViT-S/16 student)}} \\
IS16-Virchow2 & 10k & 0.508 & 0.315 & 0.421 & 0.547 & 0.244 & 0.118 & 0.235 & 0.516 & 0.238 & 0.349 \\
 & 20k & 0.528 & 0.339 & 0.418 & 0.506 & 0.252 & 0.120 & 0.232 & 0.532 & 0.237 & 0.352 \\
 & 30k & 0.534 & 0.330 & 0.429 & 0.534 & 0.247 & 0.128 & 0.214 & 0.530 & 0.237 & 0.354 \\
 & 40k & 0.537 & 0.344 & 0.426 & 0.529 & 0.252 & 0.137 & 0.223 & 0.533 & 0.239 & 0.358 \\
 & 50k & 0.540 & 0.342 & 0.430 & 0.528 & 0.254 & 0.134 & 0.225 & 0.529 & 0.240 & 0.358 \\
\addlinespace
IS16-HOpt0 & 10k & 0.506 & 0.303 & 0.440 & 0.538 & 0.242 & 0.117 & 0.242 & 0.530 & 0.242 & 0.351 \\
 & 20k & 0.517 & 0.328 & 0.449 & 0.548 & 0.244 & 0.123 & 0.241 & 0.519 & 0.251 & 0.358 \\
 & 30k & 0.535 & 0.321 & 0.451 & 0.550 & 0.256 & 0.134 & 0.227 & 0.520 & 0.253 & 0.361 \\
 & 40k & 0.538 & 0.321 & 0.452 & 0.554 & 0.249 & 0.137 & 0.227 & 0.525 & 0.254 & 0.362 \\
 & 50k & 0.538 & 0.323 & 0.453 & 0.557 & 0.248 & 0.141 & 0.234 & 0.523 & 0.253 & 0.363 \\
\addlinespace
IS16-H0mini & 10k & 0.529 & 0.327 & 0.454 & 0.549 & 0.253 & 0.121 & 0.242 & 0.541 & 0.246 & 0.362 \\
 & 20k & 0.541 & 0.325 & 0.464 & 0.563 & 0.245 & 0.136 & 0.243 & 0.543 & 0.248 & 0.367 \\
 & 30k & 0.547 & 0.339 & 0.469 & 0.576 & 0.236 & 0.136 & 0.246 & 0.545 & 0.252 & 0.372 \\
 & 40k & 0.551 & 0.335 & 0.473 & 0.578 & 0.253 & 0.142 & 0.250 & 0.547 & 0.251 & 0.376 \\
 & 50k & 0.553 & 0.340 & 0.474 & 0.584 & 0.247 & 0.140 & 0.252 & 0.550 & 0.251 & 0.377 \\
\addlinespace
IS16-UNI2h & 10k & 0.509 & 0.328 & 0.430 & 0.539 & 0.224 & 0.110 & 0.214 & 0.509 & 0.230 & 0.344 \\
 & 20k & 0.519 & 0.337 & 0.429 & 0.556 & 0.245 & 0.150 & 0.221 & 0.524 & 0.240 & 0.358 \\
 & 30k & 0.527 & 0.343 & 0.423 & 0.560 & 0.257 & 0.154 & 0.215 & 0.536 & 0.244 & 0.362 \\
 & 40k & 0.531 & 0.346 & 0.429 & 0.563 & 0.260 & 0.161 & 0.207 & 0.527 & 0.245 & 0.363 \\
 & 50k & 0.534 & 0.352 & 0.428 & 0.556 & 0.261 & 0.164 & 0.209 & 0.528 & 0.247 & 0.364 \\
\end{longtable}
\endgroup

\begingroup
\fontsize{7}{8}\selectfont
\setlength{\tabcolsep}{3pt}
\begin{longtable}{llcccccc}
\caption{Full PLISM robustness results on the 8{,}139-tile protocol across all checkpoints. PLISM is the mean of all-pairs median cosine similarity and median top-10 retrieval accuracy under scanner-only, staining-only, and paired changes. The remaining columns report median cosine similarity and median top-5 retrieval accuracy.}\label{tab:supp_plism}\\
\toprule
Model & Step & PLISM & Cosine & Top-5 & Scanner & Stain & Scan+stain \\
\midrule
\endfirsthead
\multicolumn{8}{l}{\emph{\tablename\ \thetable\ (continued)}} \\
\toprule
Model & Step & PLISM & Cosine & Top-5 & Scanner & Stain & Scan+stain \\
\midrule
\endhead
\bottomrule
\endlastfoot
\multicolumn{8}{l}{\textit{Reference encoders and baselines}} \\
Virchow2 (632M) & -- & 0.447 & 0.744 & 0.094 & 0.516 & 0.203 & 0.076 \\
UNI2-h (681M) & -- & 0.333 & 0.592 & 0.033 & 0.421 & 0.123 & 0.023 \\
H-optimus-0 (1.1B) & -- & 0.480 & 0.686 & 0.124 & 0.668 & 0.240 & 0.103 \\
Midnight-12k (1.1B) & -- & 0.337 & 0.743 & 0.060 & 0.276 & 0.111 & 0.051 \\
GPFM (303M) & -- & 0.264 & 0.594 & 0.009 & 0.253 & 0.049 & 0.006 \\
H0-mini (86M) & -- & 0.540 & 0.800 & 0.135 & 0.798 & 0.224 & 0.111 \\
kaiko ViT-S/16 (22M) & -- & 0.307 & 0.756 & 0.020 & 0.248 & 0.067 & 0.014 \\
ViT-S/16 IN21K & -- & 0.383 & 0.862 & 0.041 & 0.388 & 0.099 & 0.033 \\
\midrule
\multicolumn{8}{l}{\textit{DistillPath-KS16 (kaiko ViT-S/16 student)}} \\*
KS16-Virchow2 & 10k & 0.434 & 0.703 & 0.072 & 0.568 & 0.172 & 0.056 \\
 & 20k & 0.436 & 0.704 & 0.073 & 0.568 & 0.179 & 0.056 \\
 & 30k & 0.452 & 0.716 & 0.080 & 0.594 & 0.191 & 0.062 \\
 & 40k & 0.445 & 0.718 & 0.077 & 0.571 & 0.191 & 0.059 \\
 & 50k & 0.447 & 0.720 & 0.077 & 0.578 & 0.191 & 0.060 \\
\addlinespace
KS16-HOpt0 & 10k & 0.430 & 0.628 & 0.090 & 0.628 & 0.184 & 0.071 \\
 & 20k & 0.447 & 0.634 & 0.098 & 0.666 & 0.197 & 0.077 \\
 & 30k & 0.465 & 0.643 & 0.106 & 0.707 & 0.212 & 0.085 \\
 & 40k & 0.478 & 0.654 & 0.115 & 0.735 & 0.226 & 0.093 \\
 & 50k & 0.480 & 0.656 & 0.115 & 0.738 & 0.228 & 0.093 \\
\addlinespace
KS16-H0mini & 10k & 0.465 & 0.794 & 0.081 & 0.610 & 0.162 & 0.061 \\
 & 20k & 0.469 & 0.792 & 0.083 & 0.622 & 0.170 & 0.063 \\
 & 30k & 0.484 & 0.804 & 0.088 & 0.662 & 0.175 & 0.068 \\
 & 40k & 0.493 & 0.812 & 0.092 & 0.667 & 0.186 & 0.071 \\
 & 50k & 0.495 & 0.816 & 0.093 & 0.674 & 0.187 & 0.073 \\
\addlinespace
KS16-UNI2h & 10k & 0.432 & 0.672 & 0.069 & 0.641 & 0.162 & 0.053 \\
 & 20k & 0.456 & 0.687 & 0.075 & 0.690 & 0.181 & 0.059 \\
 & 30k & 0.469 & 0.697 & 0.081 & 0.709 & 0.193 & 0.064 \\
 & 40k & 0.483 & 0.717 & 0.086 & 0.725 & 0.206 & 0.068 \\
 & 50k & 0.484 & 0.724 & 0.086 & 0.727 & 0.206 & 0.068 \\
\midrule
\multicolumn{8}{l}{\textit{DistillPath-IS16 (ImageNet-21k ViT-S/16 student)}} \\*
IS16-Virchow2 & 10k & 0.439 & 0.713 & 0.084 & 0.588 & 0.167 & 0.069 \\
 & 20k & 0.471 & 0.735 & 0.100 & 0.632 & 0.198 & 0.083 \\
 & 30k & 0.468 & 0.739 & 0.098 & 0.634 & 0.191 & 0.080 \\
 & 40k & 0.481 & 0.742 & 0.107 & 0.658 & 0.203 & 0.088 \\
 & 50k & 0.490 & 0.746 & 0.112 & 0.675 & 0.208 & 0.093 \\
\addlinespace
IS16-HOpt0 & 10k & 0.486 & 0.687 & 0.110 & 0.771 & 0.195 & 0.090 \\
 & 20k & 0.514 & 0.696 & 0.138 & 0.813 & 0.225 & 0.115 \\
 & 30k & 0.517 & 0.706 & 0.140 & 0.806 & 0.227 & 0.117 \\
 & 40k & 0.523 & 0.705 & 0.145 & 0.819 & 0.238 & 0.121 \\
 & 50k & 0.526 & 0.710 & 0.147 & 0.811 & 0.242 & 0.123 \\
\addlinespace
IS16-H0mini & 10k & 0.517 & 0.832 & 0.112 & 0.720 & 0.199 & 0.091 \\
 & 20k & 0.522 & 0.826 & 0.119 & 0.730 & 0.204 & 0.096 \\
 & 30k & 0.537 & 0.835 & 0.129 & 0.764 & 0.213 & 0.106 \\
 & 40k & 0.540 & 0.840 & 0.130 & 0.768 & 0.213 & 0.107 \\
 & 50k & 0.543 & 0.842 & 0.133 & 0.775 & 0.217 & 0.109 \\
\addlinespace
IS16-UNI2h & 10k & 0.513 & 0.765 & 0.117 & 0.788 & 0.197 & 0.098 \\
 & 20k & 0.535 & 0.757 & 0.138 & 0.833 & 0.225 & 0.117 \\
 & 30k & 0.554 & 0.767 & 0.157 & 0.855 & 0.248 & 0.135 \\
 & 40k & 0.562 & 0.779 & 0.159 & 0.866 & 0.256 & 0.136 \\
 & 50k & 0.561 & 0.782 & 0.156 & 0.863 & 0.252 & 0.134 \\
\end{longtable}
\endgroup

\end{document}